\documentclass[]{x2lab}

\usepackage{microtype}
\usepackage{graphicx}
\usepackage{subcaption}
\usepackage{csquotes}
\usepackage{afterpage}

\usepackage{amsmath}
\usepackage{amssymb}
\usepackage{mathtools}
\usepackage{amsthm}
\usepackage{xspace}
\usepackage{colortbl}
\usepackage{multirow}
\usepackage{enumitem}
\usepackage{wrapfig}
\usepackage{nicefrac}
\usepackage{makecell}
\usepackage{multirow}
\usepackage{xcolor}

\usepackage{pifont}

\definecolor{fbApp}{HTML}{c8e7fa}
\definecolor{fbPurple3}{HTML}{f0ebf5}

\definecolor{citecolor}{HTML}{0071BC}
\definecolor{linkcolor}{HTML}{ED1C24}

\definecolor{citecolor}{HTML}{0071BC}
\definecolor{linkcolor}{HTML}{ED1C24}

\title{Igniting VLMs toward the Embodied Space}

\author[]{Andy Zhai}
\author[]{Brae Liu}
\author[]{Bruno Fang}
\author[]{Chalse Cai}
\author[]{Ellie Ma}
\author[]{Ethan Yin}
\author[*]{Hao Wang}
\author[]{Hugo Zhou}
\author[]{James Wang}
\author[]{Lights Shi}
\author[\dagger]{Lucy Liang}
\author[]{Make Wang}
\author[]{Qian Wang}
\author[]{Roy Gan}
\author[]{Ryan Yu}
\author[]{Shalfun Li}
\author[]{Starrick Liu}
\author[]{Sylas Chen}
\author[]{Vincent Chen}
\author[]{Zach Xu}

\affiliation[]{\textbf{X SQUARE ROBOT}}
\contribution[\dagger]{Project Lead}
\contribution[*]{Correspondence}

\abstract{
While foundation models show remarkable progress in language and vision, existing vision-language models (VLMs) still have limited spatial and embodiment understanding. Transferring VLMs to embodied domains reveals fundamental mismatches between modalities, pretraining distributions, and training objectives, leaving action comprehension and generation as a central bottleneck on the path to AGI.
We introduce \textbf{WALL-OSS}, an end-to-end embodied foundation model that leverages large-scale multimodal pretraining to achieve (1) embodiment-aware vision--language understanding, (2) strong language--action association, and (3) robust manipulation capability.
Our approach employs a tightly coupled architecture and multi-strategies training curriculum that enables {Unified Cross-Level CoT}—seamlessly unifying instruction reasoning, subgoal decomposition, and fine-grained action synthesis within a single differentiable framework.
Our results show that WALL-OSS attains high success on complex long-horizon manipulations, demonstrates strong instruction-following capabilities, complex understanding and reasoning, and outperforms strong baselines, thereby providing a reliable and scalable path from VLMs to embodied foundation models.
}
  
\date{September 8, 2025}

\metadata[Code]{\url{https://github.com/X-Square-Robot/wall-x}}
\metadata[Project Page]{\url{https://x2robot.com/en/research/68bc2cde8497d7f238dde690}}

\begin{document}
\maketitle

\begin{figure}[]
  \centering
  \includegraphics[width=1.\linewidth]{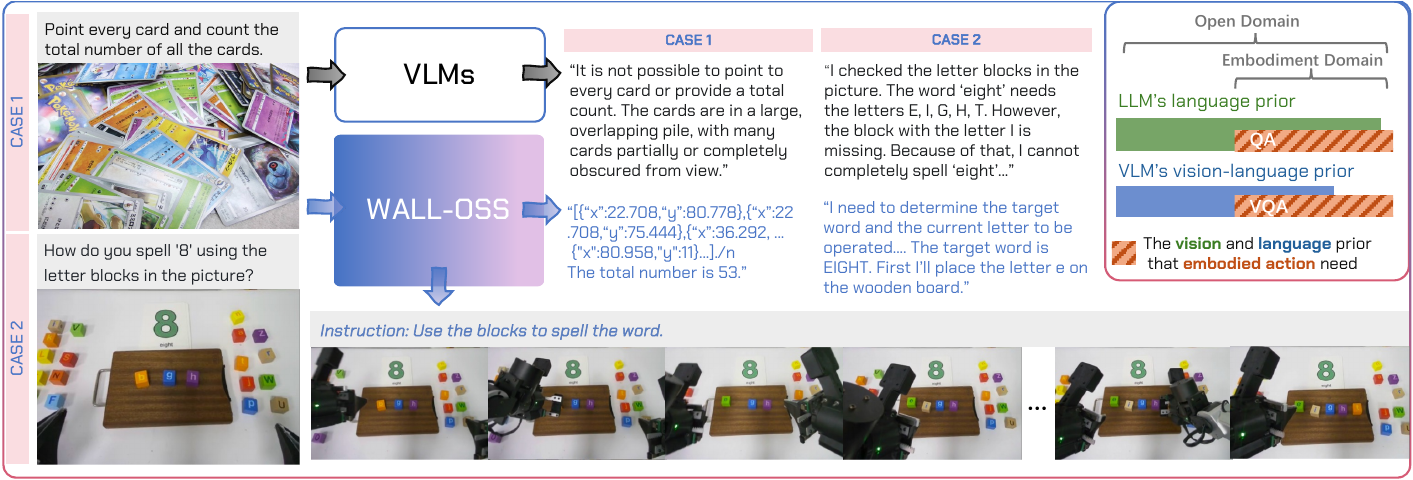}
  \caption{Current VLMs lack a sufficient understanding of space and action within embodied AI. This deficiency stems from a mismatch between the capabilities of existing pre-trained VLMs and the specific knowledge required for embodied tasks. WALL-OSS unleashes the embodied potential of VLMs, leading to enhanced embodied understanding and the ability to generate complex actions.}
  \label{fig:i2}
\end{figure}

\section{Introduction}
Foundation models in language and vision have advanced rapidly.
The trajectory of LLM progress is now mirrored in VLMs: powerful omni models such as Gemini~2.5~\cite{gemini25report} and GPT\mbox{-}5~\cite{openai_gpt5} can jointly process text and visual streams while preserving deliberate reasoning and low error across multimodal domains.
However, these systems remain largely disembodied: they neither refine themselves via feedback from interaction with the physical world nor generate executable actions. Consequently, within the embodied space, {action understanding and generation} remains a core bottleneck on the path toward AGI.

Modalities that have most successfully approached general-purpose generalization typically rely on pretraining over large-scale, diverse, and readily accessible data. Benefiting from the development of human media, the Internet hosts massive, richly distributed text, image, and video data; thus, text and 2D vision have relatively early acquired both single-modality general competence and cross-modal association. In embodied settings, however, aligned Vision--Language--Action (VLA) data are scarce and heterogeneous, making from-scratch training struggle to achieve generalization in the action modality as well as cross-modal association. This status quo has motivated transferring strong VLM backbones to the action space, e.g., OpenVLA~\cite{openvla} and $\pi_{0}$~\cite{pi0}, which model continuous actions while learning cross-modal correspondences to migrate vision and language priors (VL priors) into the action space.

A naive fine-tuning strategy---attaching an action head to a pretrained VLM and supervising it with robot trajectories---often leads to severe weight drift, thereby eroding the original VL prior. ``Knowledge insulation'' strategies preserve pretrained capabilities by minimizing perturbations to VLM parameters; however, loosely coupled architectural designs restrict the binding between semantics and control, and because VLMs themselves are deficient in the embodied space, they perform poorly when executing actions outside the VL prior. Overall, learning actions with current VLMs is challenging due to three fundamental gaps:

\textbf{Gap in modalities and data scale.} Thanks to the long-term development of computer vision, visual encoders extract features that filter high-frequency noise while retaining high-level semantics; the compressed visual tokens, together with the already highly compressed text modality, enable alignment via CLIP and related multimodal models and VLMs after training on large-scale web text--image/video pairs~\cite{radford2021clip}. In contrast, actions in the embodied space are continuous over 3D space and time; compared to the visual modality, their feature extraction lacks long-term representational study, and their high-level semantic alignment with text lacks the drive of massive data. Moreover, in embodied scenarios, multiple subtasks and concrete scene-level action descriptions are often abstracted into a single high-level instruction, further increasing cross-modal association difficulty. Pipeline-based agents (first decomposing instructions with an LLM, then invoking a skill library~\cite{ahn2022saycan,liang2023codeaspolicies}) are limited by non-differentiable interfaces and error accumulation, degrading success rates on long-horizon tasks. In addition, some methods such as 3D-VLA~\cite{3dvla} and PointVLA~\cite{pointvla} attempt to use the 3D visual modality as a bridge between 2D vision and action, but 3D data are likewise difficult to obtain, and current 3D-vision foundation models such as VGGT~\cite{vggt} and $\pi^3$~\cite{pi3} still fall short of the accuracy required for fine-grained action prediction.

\textbf{Gap in pretraining distributions.} As shown in Figure~\ref{fig:i2}, after large-scale pretraining, LLMs largely cover the needs of embodied semantics within the text modality (capable of QA). For the visual modality, however, embodied data typically involve first-person viewpoints, fisheye imaging, and self-occlusions, which differ substantially from Internet imagery. Even after large-scale pretraining, VLMs struggle to cover the needs of embodied scenarios (e.g., embodied VQA), and prior analyses have reported shortcomings in spatial reasoning, embodied-scene understanding, and progress tracking~\cite{spatial_reasoning_vlm,affordance_limits_vlm}.

\textbf{Gap in training objectives.} LLMs/VLMs use next-token likelihood on discrete sequences as their training objective, whereas action trajectories are {continuous, high-frequency} signals that are more naturally modeled with {conditional generative} objectives such as diffusion~\cite{ho2020ddpm,chi2023diffusionpolicy} or flow matching~\cite{lipman2022flow}, which learn score/velocity fields that transport a base distribution to the action distribution. However, grafting these objectives directly onto VLMs magnifies the tokenization gulf and independence assumptions, causing catastrophic degradation that weakens language--action alignment and generalization. This mirrors the challenges faced when unifying generation and understanding in current image/video settings. Some approaches (e.g., $\pi_{0}$) attempt to discretize actions at a high level under an autoregressive paradigm to align action tokens with text tokens, and then generate continuous action signals by letting action noise from diffusion/flow-matching interact with intermediate VLM representations via self-attention~\cite{pi0}. Yet such loosely coupled designs still fail to learn tight text--action bindings, leading to inadequate instruction following.

To address the above issues, we propose \textbf{WALL-OSS}, a tightly coupled VLA foundation model and training recipe. We first introduce a tightly coupled Mixture-of-Experts (MoE) architecture and training strategy that activates different experts and weights across different stages with different action modeling approaches (discrete or continuous), thereby bridging the training objective gap. We combine multimodal tasks with VQA tasks and corresponding datasets tailored to embodiment, spatial localization, and progress modeling to remedy VLM shortcomings in embodied scenarios, and implant discretized action priors into the VLM output space for coarse action supervision, thereby bridging the VLM pretraining-distribution gap. Building upon these designs, we model unified CoT forward mapping that enables the model to decompose tasks from high-level semantics to fine-grained actions, compelling the model to understand the relationships between each level in this chain, thereby bridging the modality gap. 

Our key contributions establish a principled framework for scaling VLMs to embodied intelligence. Methodologically, we demonstrate how architectural coupling and curriculum design can systematically address the fundamental misalignments between disembodied pretraining and embodied domains. Empirically, WALL-OSS achieves state-of-the-art performance across diverse manipulation benchmarks, with particularly strong results on instruction following, compositional reasoning, and long-horizon control. More broadly, our work validates a scalable paradigm for bridging the semantic-sensorimotor divide, offering both theoretical insights and practical advances toward general-purpose embodied AI systems. Given the current scarcity of embodied general foundation models, we also open-source comprehensive training code and model checkpoints to facilitate reproducible research and community development in this emerging domain.

\section{Related Work}

\bigskip
\noindent\textbf{Foundation models in language, vision and manipulation.}
Web-scale pretraining has pushed LLMs and VLMs toward stronger multimodal reasoning and grounding (e.g., Gemini~2.5 \cite{gemini25,gemini25site}, Claude~3.7 \cite{claude37}, DeepSeek-R1, Llama~3 and 4 \cite{llama31,llama4}, Qwen~2.5 \cite{qwen25}; GPT-4o \cite{gpt4o}, Qwen2.5-VL \cite{Qwen2.5-VL}, and the InternVL series \cite{internvl,internvl15,internvl25,internvl3,internvl35}). Building on these priors, VLA baselines align VLMs with robot control through encoding actions as tokens or learning continuous heads, spanning continuous models such as diffusion and flow matching (e.g. DP, Octo, $\pi_0$, $\pi_{0.5}$ \cite{diffusionpolicy, octo2024, pi0, pi05}), or discrete tokenization with compression (e.g. ACT, OpenVLA, FAST, RT-2) \cite{act,openvla,fast,rt2}. 
Our recipe follows \emph{discrete priors → continuous control}: we first implant discrete action priors into the VLM output space (Inspiration), then optimize a flow-matching head for high-frequency actions (Integration), while a tightly coupled MoE with static routing enforces strong language–action binding.

\noindent\textbf{Multimodal co-training with robot data.}
To offset the cost and coverage limits of robot trajectories, multimodal and cross-source co-training is effective. Initializing from a VLM and jointly training on web image–text, dialogs, and long videos together with multi-embodiment robot data preserves open-world language–vision competence while injecting embodied spatial semantics. Cross-embodiment and cross-environment aggregation further boosts transfer \cite{driess2023palme, octo2024, pi0, pi05, vima, vpt, rt1}. 
Our approach emphasizes embodied cognition supervision in a unified format that links instruction, reasoning, sub-tasks, and continuous actions; systematic unification of open-source action data with distribution mixing and quotas across two stages to limit drift; and a tightly coupled MoE with static routing that converts heterogeneous supervision into controllable gains on the action channel, improving robustness in unfamiliar scenes and across embodiments.

\noindent\textbf{Language reasoning and subtask decomposition.}
For long-horizon tasks, combining high-level reasoning with low-level control improves success. Unlike pipeline designs that pair a VLM planner with a separate controller (e.g., SayCan \cite{saycan} and Code-as-Policies \cite{cap}; hierarchical variants such as Hi Robot \cite{shi2025hi} and GR00T N1 \cite{gr00tn1_2025}), we adopt a single-model Uni‑CoT formulation that learns end-to-end mappings from instruction to CoT, sub-tasks, and continuous actions. This complements hierarchical and action-grammar approaches (e.g., RT‑H \cite{rt_h}), recent CoT-for-control methods \cite{cotvla2025,zawalski2024embodiedcot} and prompting frameworks \cite{pivot2024}, as well as instruction-tuning for action \cite{llarva} and hierarchical goal-conditioned policies \cite{hamster}. The model can include or skip intermediates and interleave reasoning with execution. Coupled with our two-stage curriculum and tightly coupled routing, this unified design reduces interface-induced error accumulation and strengthens progress awareness and instruction following on complex, long-horizon manipulation tasks.

\begin{figure}[h]
  \centering
  \includegraphics[width=0.80\linewidth]{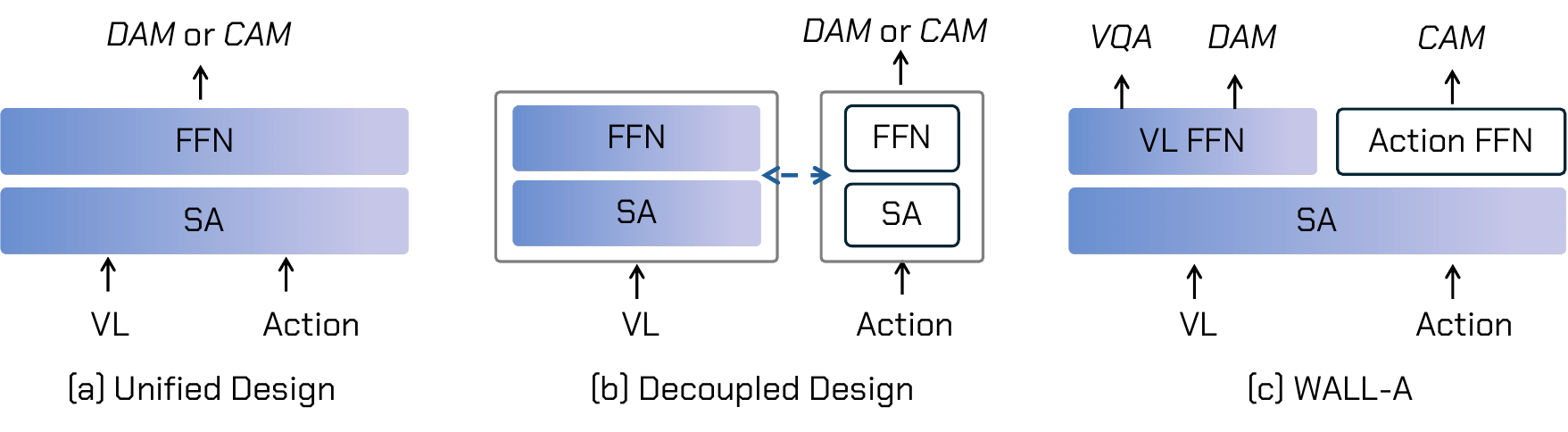}
  \caption{\textbf{Different paradigms for transferring VLMs to action modeling.} The blue parts denote the initialized weights inherited from the pretrained VLM backbone. DAM and CAM refer to Discrete Action Modeling and Continuous Action Modeling, respectively, while VL denotes Vision--Language, and SA denotes Self--Attention.}
  \label{fig:m0}
\end{figure}
\section{WALL-OSS}

In this section, we present the overall architecture and training methodology of WALL-OSS.
Our objective is to develop a transformer-based foundation model for embodied domains that not only enhances spatial understanding of embodied scenarios within the VLM but also achieves high-quality action generation.

\textbf{Model Architecture.}
Figure~\ref{fig:m0}(a) and (b) illustrate the structural paradigms of existing VLA models.
In (a), the {mixed design} directly extends the original VLM to model actions through either Discrete Action Modeling (DAM) or Continuous Action Modeling (CAM), following the next-token prediction paradigm, as exemplified by RT-2~\cite{rt2} and OpenVLA~\cite{openvla}. 
However, action supervision significantly perturbs the weight distribution of the original VLM, leading to substantial degradation in action instruction-following and generalization capabilities, resulting in overfitting to actions. 
In (b), the {decoupled design} employs a separate branch for action prediction that interacts with and extracts information from the VLM, as demonstrated in $\pi_0$~\cite{pi0}. 
The limitation is that vision and language merely serve as auxiliary signals for action generation; this loosely coupled architecture weakens the instruction-following capability of action prediction.
Our architecture addresses these limitations by leveraging a Mixture-of-Experts (MoE)~\cite{moe} design and assigning different FFNs to corresponding training tasks, thereby forming a tightly coupled cross-modal structure that effectively enhances the model's cross-modal association capabilities.

\begin{figure}[h]
  \centering
  \includegraphics[width=0.9\linewidth]{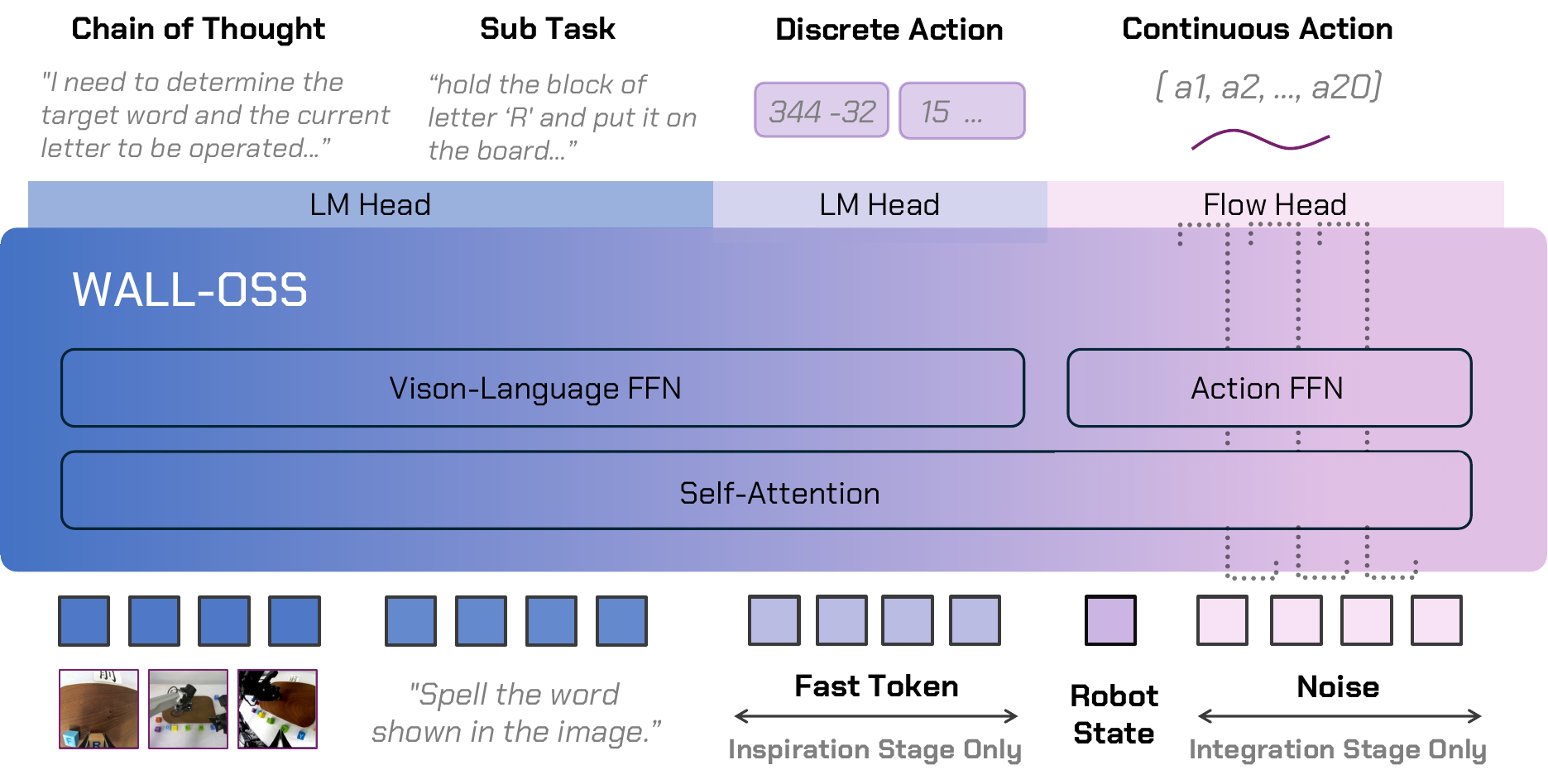}
  \caption{Architecture of \textbf{WALL-OSS}.}
  \label{fig:m1}
\end{figure}

Figure~\ref{fig:m1} presents the overall architecture of \textbf{WALL-OSS}. We adopt QwenVL2.5-3B as the main backbone. The model takes as input {vision} (egocentric and arm-mounted camera views) together with {text instructions} and produces different outputs depending on the training phase, while remaining conditioned on the same multimodal inputs throughout. Let $\mathbf{c}=(\text{vision},\text{instruction})$ and $\mathbf{h}=F_{\theta}(\mathbf{c})$ denote the input pair and its VLM encoding, respectively, where $F_{\theta}$ represents the VLM backbone with parameters $\theta$.

\begin{figure}[h]
  \centering
  \includegraphics[width=0.8\linewidth]{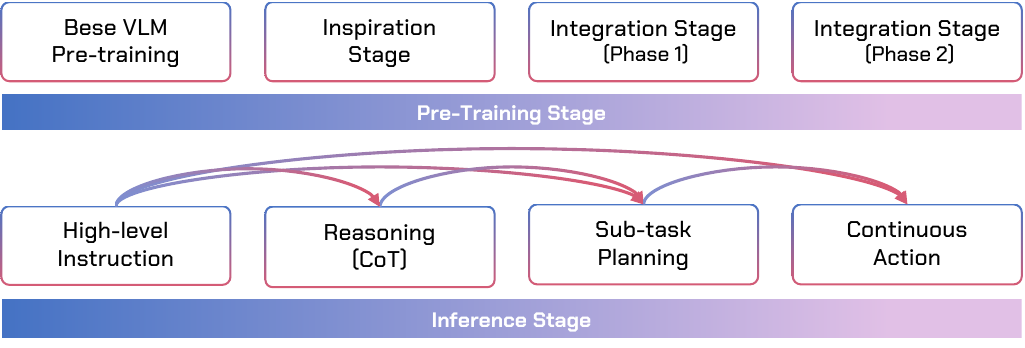}
  \caption{Overview of training and inference pipeline.}
  \label{fig:m2}
\end{figure}

Following the pretraining of VLMs, our pretraining process comprises two main components: \textbf{Inspiration} of the VLM and \textbf{Integration} of the three modalities (Vision-Language-Action), as shown in Figure~\ref{fig:m2}.

\textbf{Inspiration Stage.} We begin by reusing the original feed-forward networks (FFNs) of the pretrained VLM and augmenting them with embodied VQA to strengthen spatial reasoning in robotic environments. The training objectives encompass masked language modeling, image/video--text contrastive learning, instruction following, and temporal order/causality modeling to build strong grounded VL priors. In parallel, we introduce a discrete action objective in which textual tokens are aligned with discrete action tokens obtained via \textsc{FAST} tokenization (cf.\ $\pi_0$-FAST~\cite{pi0}):
\[
\underbrace{\mathbf{z}_{1:K}=\textsc{FAST}(\mathbf{a})}_{\text{DCT}\ \rightarrow\ \text{Quant}\ \rightarrow\ \text{BPE}},
\qquad
\mathcal{L}_{\text{Inspiration}}
=\lambda_{\text{VQA}}\!\sum_{t}-\log p_{\theta}(\tau_t\!\mid\!\tau_{<t},\mathbf{c})
+\lambda_{\text{D}}\!\sum_{k}-\log p_{\theta}(z_k\!\mid\!z_{<k},\mathbf{c}).
\]
where $\mathbf{a}$ represents the continuous action trajectory, $\mathbf{z}_{1:K}$ are the discrete action tokens obtained through FAST tokenization, $\tau_t$ denotes the $t$-th text token, and $\lambda_{\text{VQA}}$ and $\lambda_{\text{D}}$ are weighting hyperparameters for VQA and discrete action objectives, respectively.
This phase equips the VLM with coarse, semantically grounded action awareness; its outputs include chain-of-thought reasoning, sub-task predictions, and discrete \textsc{FAST} action tokens. The embodied VQA enhances the VLM's spatial understanding capabilities in embodied environments, while the \textsc{FAST} token prediction provides coarse action comprehension. This {inspires} the initial VLM with fundamental embodied reasoning and action awareness.

\textbf{Integration Stage.} Building on these priors, we replace discrete action prediction with continuous action modeling via flow matching, and partition this stage into two phases: (1) freeze the VLM and train only the flow head under the Action FFN, and (2) unfreeze the VLM for joint optimization.
For the integration stage, vision, language, and action representations interact through attention, while a static router directs action-centric features to an {Action FFN} and vision--language features to a {Vision--Language FFN}, instead of a learned softmax/top-k router. 
In the first phase, we form noisy samples and regress the velocity field
\[
x_t=(1-\rho(t))\,x_0+\rho(t)\,\epsilon,\qquad
\mathcal{L}_{\text{Integration}}
=\lambda_{\text{C}}\ \mathbb{E}\!\left[w(t)\,\big\|v_{\phi}(x_t,\mathbf{h},t)-(\epsilon-x_0)\big\|_2^2\right],
\]
where $x_0$ represents the clean action sample, $x_t$ is the noisy sample at time $t$, $\epsilon$ is Gaussian noise, $\rho(t)$ is a noise schedule function, $v_{\phi}$ is the velocity field network with parameters $\phi$, $w(t)$ is a weighting function, and $\lambda_{\text{C}}$ is the continuous action loss weight.
The supervision from the Inspiration stage stabilizes cross-modal attention here, preserving VL priors and providing reliable initialization for continuous action.

For the second phase, we jointly optimize both modules using the same flow-matching objective with gradient routing $\partial \mathcal{L}_{\text{Integration}}/\partial \theta \neq 0$ and $\partial \mathcal{L}_{\text{Integration}}/\partial \phi \neq 0$ (unfreeze the VLM and train jointly with the action branch). 
This {integration} enables fine-grained action output within the tightly coupled architecture, compelling the model to integrate multimodal information and efficiently accomplish multimodal tasks, thereby achieving alignment across different modalities.


\textbf{Unified Cross-Level CoT.}
Our approach generalizes the notion of chain-of-thought reasoning beyond traditional narrow-sense CoT (step-by-step textual reasoning in LLMs) to encompass a broad-sense CoT that spans the entire semantic-to-sensorimotor spectrum: $\text{instruction}\rightarrow\text{reasoning (CoT)}\rightarrow\text{subtask plan}\rightarrow\text{continuous actions}$. This unified formulation enables forward arbitrary mapping across hierarchical abstraction levels, allowing the model to seamlessly transition between high-level deliberation and low-level execution within a single differentiable framework.

Existing approaches often decompose these levels into pipeline-style or multi-module systems, where an instruction is first passed to a planner to generate a high-level plan, which is then executed by a controller, such as Hi Robot~\cite{shi2025hi} and GR00T N1~\cite{gr00tn1_2025}. While these paradigms can substantially reduce the difficulty of training and execution in the short term, they introduce non-differentiable interfaces, limit system performance by the capacity of each module, and tend to accumulate errors across stages.  

In contrast, \textbf{WALL-OSS} employs a {single}, end-to-end model that jointly learns mappings across abstraction levels, supporting flexible forward arbitrary mapping where the model can adaptively leverage full intermediate reasoning steps or directly map instructions to actions based on task complexity and contextual demands.  

Formally, let $v$ denote visual input, $x$ the language instruction, $c$ an optional chain-of-thought, and $a_{1:T}\in\mathbb{R}^{T\times d}$ the target action trajectory of length $T$ with dimension $d$. Uni-CoT trains a unified predictor $F_\theta$ with a path-drop objective that allows the model to condition on or bypass intermediate reasoning flexibly:
\[
\min_{\theta}\ \mathbb{E}_{(v,x,c,a)} 
\Big[\;\ell_{\text{act}}\!\big(F_\theta(v,x,c),a_{1:T}\big)\;+\;\lambda\,\ell_{\text{VQA}}(H_\theta(v,x),y)\Big],
\]
where $H_\theta$ is an embodied-aware VQA head with supervision $y$, $\ell_{\text{act}}$ is the action prediction loss, $\ell_{\text{VQA}}$ is the visual question answering loss, and $\lambda$ is a balancing hyperparameter. This formulation enables both full-chain and direct $x\!\to\!a$ training within a single model, ensuring end-to-end differentiability while retaining flexibility in reasoning.

By covering mappings across semantic levels, Uni-CoT (i) enhances spatial understanding in VLMs through embodied VQA, leading to stronger grounding and action prediction, and (ii) significantly improves long-horizon task success rates and instruction following. Moreover, as shown in the lower part of Figure~\ref{fig:m2}, during inference, Uni-CoT can adaptively decide whether to invoke CoT/subtask decomposition, and even interleave reasoning with execution—emitting actions for completed subtasks while continuing to reason—thus enabling flexible, asynchronous control for real-time human–robot interactions.

\section{Data Composition}

\begin{figure*}[t]
  \centering
  \includegraphics[width=0.98\linewidth]{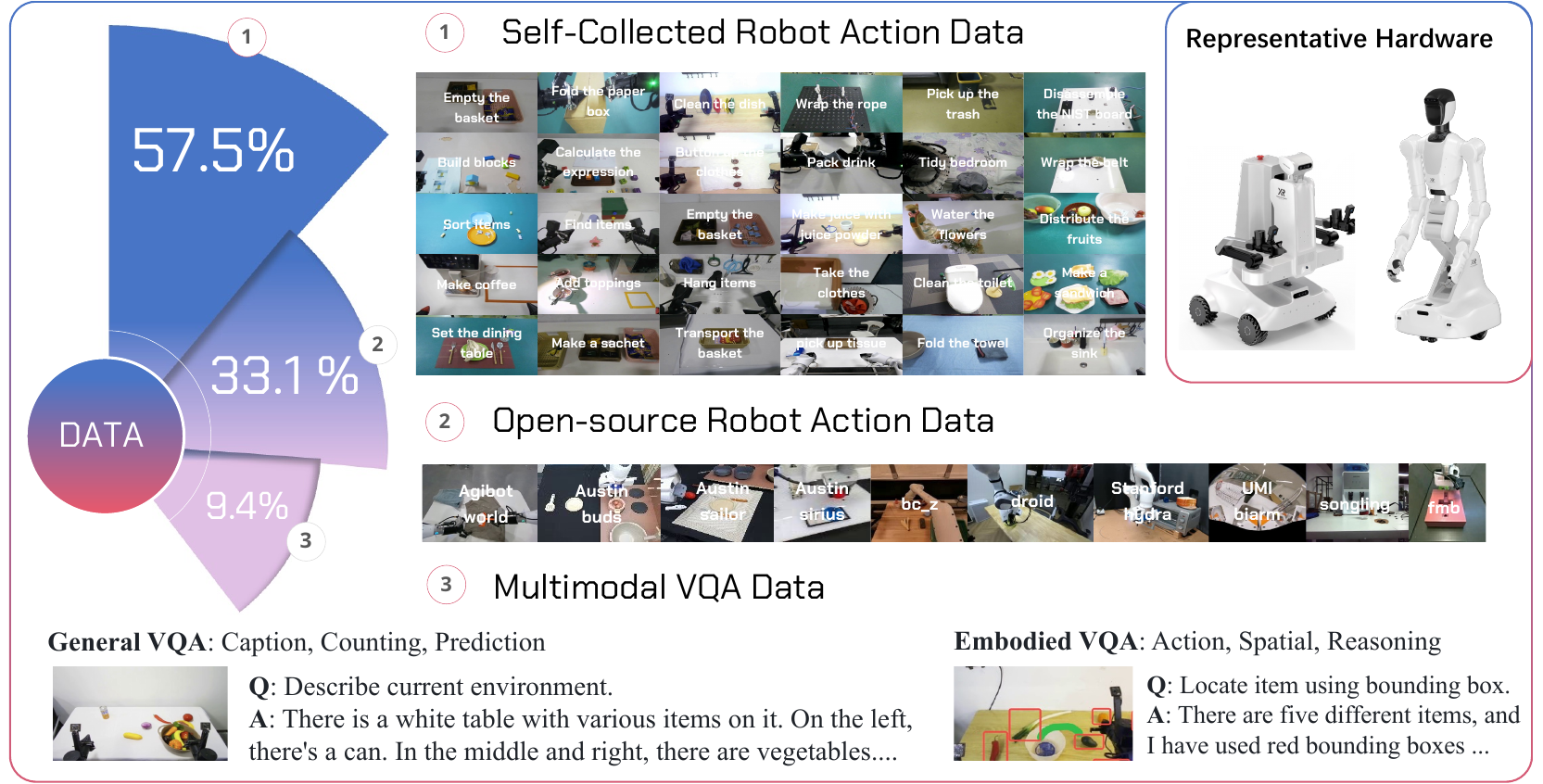}
  \vspace{-6pt}
  \caption{Overview of the multisource dataset. Left: composition across three sources (self-collected actions, open-source actions, and multimodal VQA). Middle (top to bottom): example images from self-collected actions, open-source actions, and multimodal VQA. Top-right: our representative robot hardware.}
  \label{fig:dataset_overview}
\end{figure*}

We construct an embodiment-centric, multisource dataset to address the lack of large-scale, aligned VLA supervision and the spatial understanding gaps of current VLMs. The corpus exceeds 10,000 hours and comprises three complementary components: (1) self-collected robot action data for high quality and task complexity, (2) open-source action data for cross-morphology and cross-environment generalization, and (3) multimodal VQA data to preserve and strengthen language--vision ability while providing additional supervision for spatial--temporal and reasoning, as shown in Figure~\ref{fig:dataset_overview}.

To align with our two-stage training recipe (Inspiration and Integration), we functionally orchestrate the data sources: (1) Inspiration focuses on embodied VQA, instruction following, and discrete action priors via \textsc{FAST} to inject coarse action awareness into the VLM while improving spatial reasoning; (2) Integration focuses on high-frequency continuous control with flow matching over real and unified open-source trajectories, first training the action branch and then jointly optimizing with the VLM to tighten language--vision--action alignment and mitigate forgetting.

\subsection{Self-collected robot action data}
\noindent\textbf{Platforms and scenes.} We target real-world daily manipulation across desktop arms, mobile stands, wheeled bi-arm systems, and wheeled humanoids, with egocentric/exocentric and arm-mounted cameras. Scenes span kitchen cleaning, dressing/organizing, mobile pick-and-place, and assembly tasks.

\noindent\textbf{Task spectrum.} We include two core families: (1) short-horizon manipulation under explicit instructions emphasizing precision and generalization; and (2) long-horizon reasoning tasks with goal clarity but implicit procedures, requiring task decomposition, progress tracking, and real-time decisions to complete multi-step routines.

\noindent\textbf{Annotation and quality control.} We build a multi-model pipeline for fine-grained step annotations with human spot checks, enabling CoT-style stage supervision directly on trajectories. We enforce multi-sensor timestamp synchronization, outlier filtering, low-quality/idle-frame removal, rule-based validation plus manual audits, and automatic augmentations (illumination/background). To balance with open-source actions and multimodal corpora, we select a representative self-collected subset for WALL-OSS training.

\subsection{Open-source action data}
\noindent\textbf{Coverage and sources.} We aggregate diverse datasets to complement sensor forms, operation styles, and environmental variation, with representative sources including DROID, BC-Z, BRIDGE, and others (see Table~\ref{tab:datasets_overview} for a complete list).

\noindent\textbf{Standardization.} To enable stable joint pretraining across heterogeneous sources, we enforce a unified specification: (1) coordinate frames and units (meters for position; radians for angles); (2) morphology normalization via a maximally expressive DoF template with masking/placeholders for missing joints across single-/dual-arm, wheeled, and humanoid platforms; (3) perception alignment by unifying intrinsics/extrinsics and timestamps, resampling frame rate/resolution, and channel-aligning multi-view video; and (4) action time-base normalization by standardizing control frequencies and resampling/interpolating trajectories to the flow-matching grid. This protocol reduces cross-source friction and improves optimization stability.

\subsection{Multimodal VQA: general and embodied}
We adopt two VQA streams: a general, perception-centric stream to maintain VLM capability, and an embodied stream that targets spatial--temporal and task reasoning for manipulation.

\noindent\textbf{General VQA for VLM maintenance.} We combine open-source VQA datasets (e.g., CapsFusion, RoboPoint, Robo2VLM, COCO; see Table~\ref{tab:datasets_overview}) and general VQA annotations over self-collected trajectories to maintain and enhance language--vision competence and instruction following. These signals are weakly related to manipulation and primarily regularize the vision--language backbone rather than the action head.

\noindent\textbf{Embodied VQA for spatial--temporal and reasoning.} We build an automatic generation pipeline over self-collected trajectories to supply supervision that targets spatial understanding and task reasoning: Action Planning VQA (subgoal reasoning), Spatial \& Temporal QA (localization, ordering, progress), Perception VQA (attributes), and Cognition \& Affordance (interactability). Labels use a unified format for cross-task sharing, e.g., \texttt{<box>[x1,y1,x2,y2]</box>}, \texttt{<point>[x,y]</point>}, optional \texttt{<mask>} encodings, and natural-language QA/description. These signals explicitly link the chain ``instruction \(\rightarrow\) CoT \(\rightarrow\) sub-tasks \(\rightarrow\) continuous actions'' and directly serve Uni-CoT and our tightly coupled MoE architecture.

\begin{table}[t]
  \centering
  \vspace{-2pt}
  {\footnotesize
  \setlength{\tabcolsep}{6pt}
  \resizebox{\linewidth}{!}{
  \begin{tabular}{l p{0.78\linewidth}}
    \hline
    \textbf{Category} & \textbf{Datasets} \\
    \hline
    Open-source action data & Agibotworld~\cite{bu2025agibot}; Droid~\cite{khazatsky2024droid}; Bc\_z~\cite{jang2022bc}; RH20T~\cite{fang2023rh20t}; Furniture\_bench~\cite{heo2023furniturebench}; Fractal~\cite{gaydon2024fractal}; Bridge\_data\_v2~\cite{walke2023bridgedata}; DobbE~\cite{shafiullah2023bringing}; UMI-biarm~\cite{liu2024fastumi}; Utaustin\_mutex~\cite{shah2023mutex}; Stanford\_hydra~\cite{belkhale2023hydra}; Austin\_sailor~\cite{nasiriany2022learning}; Fmb~\cite{luo2025fmb}; Austin\_sirius~\cite{liu2022robot}; Taco\_play~\cite{rosete2023latent,mees2022grounding}; Stanford\_kuka\_multimodal~\cite{lee2019making}; Berkeley\_autolab\_ur5~\cite{BerkeleyUR5Website}; Jaco\_play~\cite{dass2023jacoplay}; Viola~\cite{zhu2022viola}; Berkeley\_fanuc\_manipulation~\cite{zhu2023fanuc}; Berkeley\_cable\_routing~\cite{luo2024multistage}; Austin\_buds~\cite{zhu2022bottom}; Dlr\_edan\_shared\_control~\cite{vogel2020edan,quere2020shared}; Nyu\_rot~\cite{haldar2023watch} \\
    \hline
    Open-source image--text data & CapsFusion~\cite{CapsFusion}; Cambrian~\cite{Cambrian}; PixMo~\cite{PixMo}; RoboPoint~\cite{RoboPoint}; Robo2VLM~\cite{Robo2VLM}; COCO~\cite{COCO}; VQAv2~\cite{VQAv2}; VQASynth-SpaceLLaVA~\cite{vqasynth_spacellava}; SpaceThinker~\cite{vqasynth_spacellava}; OpenSpaces\_MC\_R1~\cite{vqasynth_spacellava}; OpenSpaces~\cite{vqasynth_spacellava}; SpaceOm~\cite{vqasynth_spacellava} \\
    \hline
  \end{tabular}
  }
  }
  \caption{Summary of open-source datasets used in our corpus. Names are shown explicitly; citations appended where available.}
  \label{tab:datasets_overview}
\end{table}

\subsection{Splits, sampling, and scale}
We stratify by scene/object/task/morphology to form cross-environment and cross-morphology validation/test sets. Long-horizon and rare skills receive temperature-controlled resampling and hard-example upsampling to strengthen long-range dependencies and progress modeling. During both training stages, we mix sources with quota control to preserve VLM capabilities and improve continuous control jointly. We apply light domain randomization and occlusion perturbations to the visual stream for robustness. The overall corpus exceeds tens of thousands of hours; source-wise proportions across duration, scenes, and morphologies, as well as task breakdowns. We provide qualitative visualizations and show in ablations that embodied VQA plus discrete action priors significantly improve subsequent flow-matching control quality and instruction adherence, especially in long-horizon tasks.
\section{Experiments}

\begin{figure*}[]
  \centering
  \includegraphics[width=\textwidth, height=1.0\textheight, keepaspectratio]{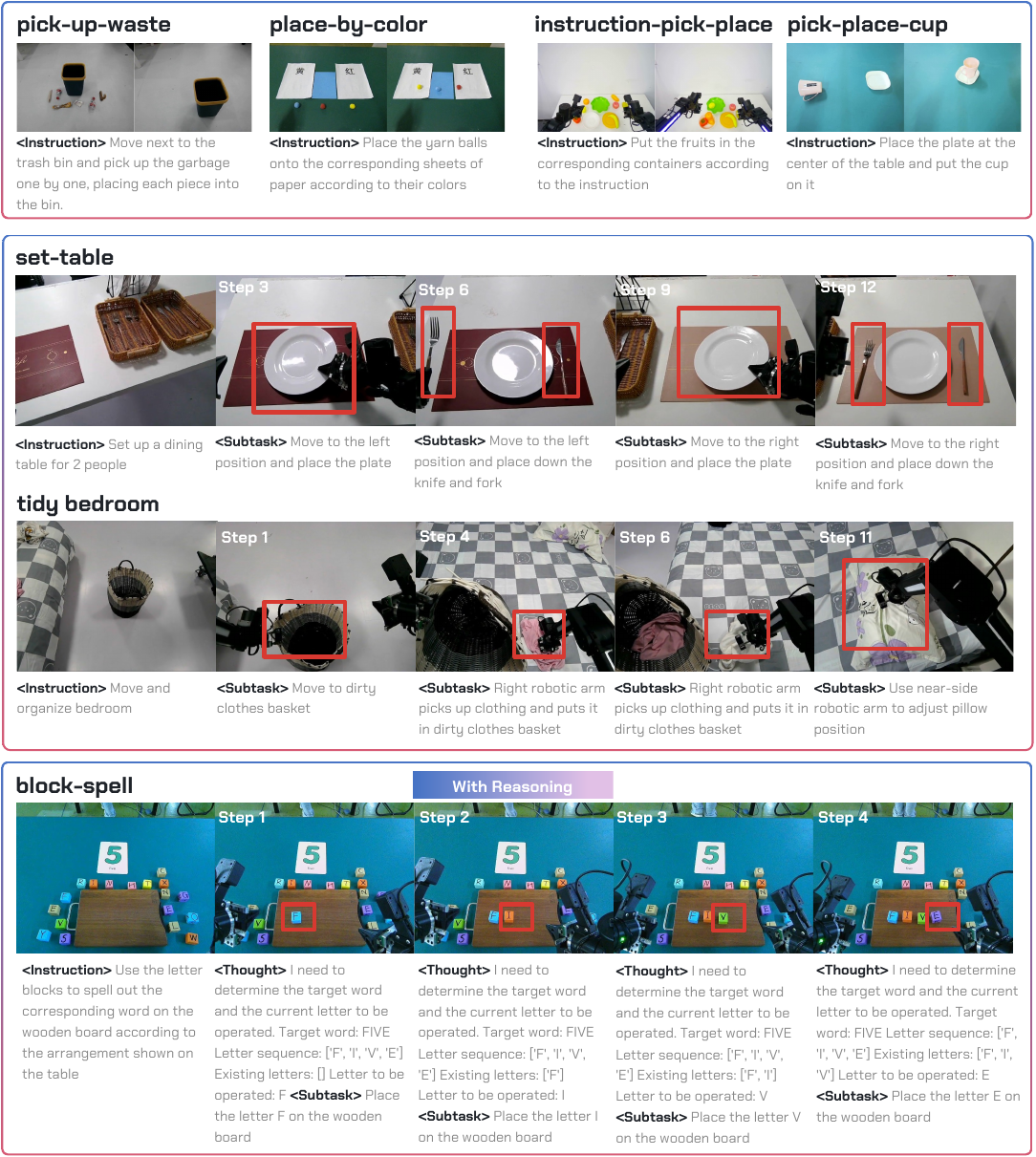}
  \caption{Overview of evaluation tasks used during fine-tuning. Top: single-instruction tasks (Pick-Up-Waste, Place-by-Color, Instruction-Pick-Place, Pick-Place-Cup). Bottom: long-horizon and reasoning tasks (Set-Table, Block-Spell, Tidy-Bedroom).}
  \label{fig:eval_tasks}
\end{figure*}

\subsection{Setup}

To comprehensively evaluate the capabilities of our VLA model, we designed an evaluation suite comprising an Embodied VQA benchmark and six robotic manipulation tasks. This suite is centered around three core dimensions: 1) language instruction understanding, reasoning, and generalization; 2) planning and execution for long-horizon, multi-stage tasks; and 3) action accuracy and robustness.

\subsubsection{Evaluation Tasks}

\textbf{Embodied Visual Question Answering}
To evaluate the model's understanding of embodied scenes, we developed an Embodied VQA benchmark. Based on video frames randomly sampled from our action dataset, this benchmark requires the model to perform three tasks: (1) Scene Captioning, which generates a textual description for a given image; (2) Object Grounding, which locates the 2D coordinates of an object specified by a textual description; and (3) Action Planning, which plans the next action in natural language given an image and a high-level instruction. We compared the performance of our pre-trained WALL-OSS model against the original base VLM (qwen2.5-vl-3b) on this benchmark via manual evaluation.

\textbf{Robotic Manipulation Tasks}
We designed six manipulation tasks to assess the model's performance across our core dimensions. Among these, {set-table}, {tidy-bedroom}, and {place-by-color} are novel tasks unseen during pre-training, designed to evaluate the model's adaptation capability on novel tasks. The tasks are grouped by the primary capability they target:

For instruction following and reasoning:

\noindent\textbf{Instruction-pick-place:} The model is required to follow diverse natural language instructions—such as descriptive commands—to pick up a specified object and place it at a designated target location. This task is excluded from fine-tuning and used to evaluate zero-shot generalization.

\noindent\textbf{Place-by-color:} The model must place yarn balls onto paper of the corresponding color. This task includes two conditions: (1) direct visual color matching (e.g., placing a red yarn ball on red paper), and (2) text-based reasoning (e.g., placing the ball on white paper printed with the word "red"). 500 episodes are used for fine-tuning.

\noindent\textbf{Block-spell:} This task requires the model to spell out answers using toy blocks. It evaluates the model's reasoning ability, as the answer must be fully inferred before any spelling action can be taken. The prompts fall into two categories: (1) identifying and spelling the name of an object depicted on an image card, and (2) solving a simple arithmetic expression and spelling the resulting answer. 1600 episodes are used for fine-tuning.

For long-horizon planning:

\noindent\textbf{Set-table:} The model must set two table placements by taking a plate, a knife, and a fork from the center of the table and arranging them correctly for each setting (plate in the middle, cutlery on either side). 1500 episodes are used for fine-tuning.

\noindent\textbf{Tidy-bedroom:} The model needs to tidy a bedroom by collecting dirty clothes from the bed, placing them into a laundry basket, and arranging the pillows correctly. 1000 episodes are used for fine-tuning.

For action accuracy and robustness:

\noindent\textbf{Collect-waste:} The model must navigate to a trash can, collect scattered waste from the surrounding area, and perform accurate disposal. This task involves dynamic object interaction and spatial reasoning under mobility constraints. 900 episodes are used for fine-tuning.

\noindent\textbf{Pick-place-cup:} The model must place a cup onto a plate centered on the table. This task challenges the model with varied initial states—such as an inverted cup, a misoriented handle, or a misplaced plate—that require a dynamic sequence of manipulations like reorienting the cup and repositioning the plate before the final placement. 500 episodes are used for fine-tuning.

\subsubsection{Baselines and Fine-tuning Configuration}

We compare WALL-OSS against two representative baselines that capture distinct paradigms in VLA learning: $\pi_0$~\cite{pi0}, which leverages pre-trained VLMs with flow matching for action generation, and Diffusion-Policy~\cite{diffusionpolicy}, which employs conditional denoising diffusion for visuomotor control without VLM initialization.

To ensure comprehensive evaluation, we implement two instruction paradigms for baseline models: \textit{Flat} training with high-level task instructions, and \textit{GPT4-Subtask} training using human-annotated subtask decompositions with GPT-4 inference-time generation. This design enables us to isolate the contribution of task decomposition from architectural innovations.

For fair backbone comparison, WALL-OSS and $\pi_0$ leverage pre-trained VLM weights, while Diffusion-Policy trains from scratch due to its smaller capacity. All models receive identical task supervision, with WALL-OSS uniquely supporting joint action-VQA training through interleaved sampling (1:15 ratio for non-subtask VQA, 1:100 for subtask samples).

\subsubsection{Evaluation Protocol}

To ensure objectivity and fairness in evaluation, we implemented a rigorous blind third-party testing protocol. The model trainers provided a detailed testing document that specified the environment setup, initial state for each task, and a standardized scoring rubric. Subsequently, third-party evaluators, who were not involved in the model's development or training, conducted the assessments. They were blind to the version of the model being tested and strictly adhered to the document provided to perform evaluations and record results. This protocol minimizes subjective bias and ensures the reliability of our evaluation outcomes.

\subsection{Results and Analysis}

We present our experimental results, starting with an evaluation of the model's foundational scene understanding and progressing to its performance on increasingly complex manipulation tasks.

\subsubsection{Pre-training Enhances Embodied Scene Understanding}

We first verify that our pre-training strategy not only preserves the VLM's visual-language alignment but also enhances its deep understanding of embodied scenes before assessing action generation capabilities. As shown in Table~\ref{tab:vqa}, we compare WALL-OSS with its original base VLM (qwen2.5-vl-3b) on our self-collected Embodied VQA benchmark. In the Object Grounding task, WALL-OSS achieves significantly superior performance; the base VLM, unfamiliar with embodied manipulation scenes that are rare in its pre-training data, is often misled by elements such as the robot arm, leading to localization failures. A similar trend emerges in the Scene Captioning task, where the base VLM often generates hallucinations irrelevant to the scene, whereas WALL-OSS accurately describes the manipulation actions of the robot arm. Finally, in Action Planning, both models are prone to misjudging the current task stage, yet WALL-OSS provides more specific and on-topic responses than the base model. These findings collectively demonstrate that our pre-training strategy successfully injects robot-centric scene knowledge into the VLM, laying a critical foundation for its performance on subsequent manipulation tasks.

\begin{figure*}[b]
  \centering
  \includegraphics[width=0.98\linewidth]{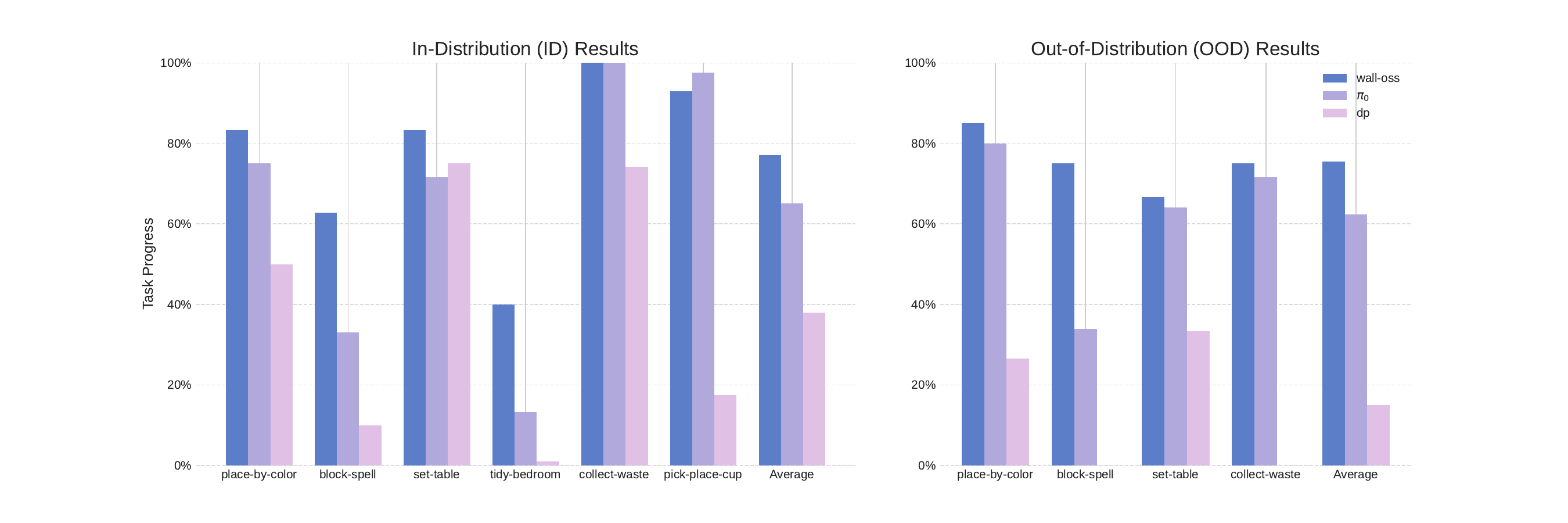}
  \vspace{-6pt}
  \caption{Performance comparison with state-of-the-art policies across all evaluation tasks. Left: In-distribution (ID) performance comparison. Right: Out-of-distribution (OOD) performance}
  \label{fig:evaluation}
\end{figure*}

\begin{table}[]
\centering
\small
\setlength{\tabcolsep}{4pt}
\begin{tabular}{lccc}
\hline
\textbf{Model}         & \textbf{Object Grounding} & \textbf{Scene Captioning} & \textbf{Action Planning} \\ \hline
Qwen2.5-VL-3B & 46.1\%           & 57.7\%           & 59.8\%         \\
WALL-OSS      & 91.6\%           & 87.6\%           & 69.0\%         \\ \hline
\end{tabular}
\caption{Embodied VQA benchmark results comparing WALL-OSS with the baseline Qwen2.5-VL-3B model across three evaluation categories. }
\label{tab:vqa}
\end{table}

\subsubsection{Zero-Shot Instruction Following Emerges After Pre-Training}
Having established the model's enhanced scene understanding, we next assess whether \textbf{WALL-OSS} can follow novel instructions without task-specific fine-tuning. This analysis focuses on verifying the model's ability to generalize to unseen task formulations by leveraging its pre-trained capabilities.

As illustrated in Figure~\ref{fig:evaluation}, we evaluate the model on a pick-and-place task under two instruction types: (1) those involving objects and containers encountered during pre-training, and (2) those featuring entirely novel items. The model achieves an average task progress of 85\% on seen-object instructions and maintains a performance of 61\% on novel-object instructions.
Most failure cases are attributed to minor pose inaccuracies when selecting grasping or placement positions for unfamiliar objects, rather than semantic misinterpretation of the instructed targets. 

These results demonstrate that WALL-OSS can accurately interpret and execute manipulation instructions in a zero-shot setting. This capability is enabled by its multimodal pre-training, which preserves the VLM’s inherent reasoning abilities and supports robust generalization to novel tasks.

\subsubsection{Pre-training Boosts Action Accuracy and Generalization}

In addition to instruction comprehension and reasoning, effective embodied intelligence requires precise and robust action execution. We therefore evaluate the impact of pre-training on raw action accuracy and generalization through two tasks: Collect-Waste and Pick-Place-Cup.

The results show that with sufficient demonstration data (Collect-Waste, 1000 demonstrations), the pre-trained models—WALL-OSS and $\pi_0$—achieve a 100\% success rate on the in-distribution (ID) task, while the non-pre-trained DP model reaches only 80\%. However, when the task complexity increases and data availability is reduced (Pick-Place-Cup, 500 demonstrations), the performance gap widens substantially: the pre-trained models maintain a success rate above 90\%, whereas DP drops below 20\%. In the out-of-distribution (OOD) generalization test—performing Collect-Waste in a novel environment—DP's success rate falls from 80\% to 0\%, completely failing to complete the task. In contrast, both WALL-OSS and $\pi_0$ maintain success rates above 80\%, demonstrating strong robustness.

These results clearly indicate that pre-training is essential for VLA models, significantly improving action accuracy, data efficiency (i.e., few-shot learning), and generalization to novel environments.

\subsubsection{Subtask Generation Improves Performance on Long-Horizon Tasks}

Long-horizon tasks present significant challenges for embodied intelligence models due to their inherent complexity, delayed supervision, and the compounding effect of execution errors across stages. To address this, we investigate whether dynamically decomposing high-level goals into intermediate subtasks—generated by a language model—can improve task success rates and execution stability.
We design two long-horizon tasks \textbf{Set-Table} and \textbf{Tidy-Bedroom} that are not included in the pre-training data to evaluate the generalization performance of WALL-OSS.
Each task comprises more than five sequential stages, with average execution times exceeding three minutes for Set-Table and five minutes for Tidy-Bedroom. 

During fine-tuning, WALL-OSS is jointly trained on action generation and subtask prediction.
Despite having only 1\% of the training data labeled with subtasks, the model consistently learns to generate high-quality subtask instructions.
At inference time, WALL-OSS first generates a subtask instruction and then uses it to guide action generation. As shown in Figure~\ref{fig:evaluation}, WALL-OSS significantly outperforms both $\pi_0$ and Diffusion-Policy across the Set-Table and Tidy-Bedroom tasks.
Qualitative analysis indicates that while all models exhibit comparable performance on low-level actions (e.g., picking up a plate), the baselines frequently suffer from stage confusion due to the absence of explicit subtask guidance.
This manifests as repetitive or misplaced actions—such as repeatedly placing cutlery at the same location or failing to set the remaining parts of the table—because the model lacks a clear notion of task progression.
Moreover, when multiple plausible actions exist under similar visual observation, baseline models tend to struggle with disambiguation and often make inconsistent or suboptimal choices, further compounding execution errors across stages.
In the more spatially complex Tidy-Bedroom task, these issues become more pronounced: baseline models often get stuck in repetitive, ineffective behaviors—especially when target objects (e.g., a missed piece of clothing) fall outside the current field of view.

In contrast, WALL-OSS leverages its multimodal reasoning capabilities to generate contextual subtask cues, enabling it to disambiguate between plausible options, maintain coherent task progression, and reliably complete all stages.

\subsubsection{Chain-of-Thought (CoT) Empowers Embodied Planning and Reasoning}

Complex embodied tasks often require reasoning over spatial relationships, object attributes, and the consequences of sequential actions. While existing VLA models directly map instructions to actions, they often struggle with tasks that demand intermediate decision-making and commonsense reasoning. To address this, we investigate the role of CoT prompting in enabling WALL-OSS to plan and execute multi-step behaviors through explicit intermediate reasoning.

Inspired by the success of large language models (LLMs), we explore transferring CoT reasoning capabilities to VLA models. To evaluate the integration of reasoning and action execution, we design two tasks—Place-by-Color and Block-Spell—that require intermediate logical inference. During fine-tuning, WALL-OSS is trained to jointly generate a CoT reasoning trace and a sub-instruction, using supervision on only 1\% of the training frames.
At inference time, the model autonomously generates both a CoT reasoning trace and a sub-instruction, which are then used as conditional inputs for action generation. In the Place-by-Color task, we observe that CoT provides minimal benefit when the goal involves direct visual matching (e.g., matching a red yarn ball to red paper).

However, in the more challenging condition requiring text-based reasoning—such as placing the yarn ball on paper printed with the word “red”—WALL-OSS achieves a significantly higher task completion rate than all baselines. This demonstrates that direct-action models are limited to visually intuitive tasks, while incorporating CoT is essential for solving instruction-conditioned tasks that require intermediate reasoning.

In the Block-Spell task, baseline models under the flat setting achieve near-zero task progress, reaffirming that the absence of task decomposition is detrimental for reasoning-intensive tasks. Consequently, in Figure~\ref{fig:evaluation}, we report baseline results only under the GPT-subtask setting.
As shown, WALL-OSS significantly outperforms the GPT-subtask baseline in both in-distribution (ID) and out-of-distribution (OOD) scenarios. Our analysis suggests that while GPT-4 can often infer correct sub-instructions, its response lacks the timeliness and contextual adaptability exhibited by WALL-OSS—particularly in complex settings such as obstructed egocentric views. Moreover, baseline models exhibit a disconnect between the high-level policy generated by GPT and the low-level action execution module, leading to poor alignment and degraded instruction following performance.

\subsubsection{Multi-modal Co-training Enhances Fine-grained Instruction Following}

Beyond general instruction following, many real-world tasks require precise grounding of fine-grained visual and linguistic cues. We therefore examine the effectiveness of multi-modal co-training in improving the model’s ability to interpret and execute detailed, object-specific instructions.

As shown in Table~\ref{tab:Instruction_following_acc}, WALL-OSS achieves significantly higher accuracy than $\pi_0$ on the Block-Spell task, particularly in identifying and placing the correct sequence of letter blocks.
In the same task, we observed that $\pi_0$ performed substantially worse than WALL-OSS in identifying and manipulating specific letter blocks. To investigate this gap, we conducted an ablation study comparing three fine-tuning configurations:

\begin{itemize}
    \item  WALL-OSS (Multi-modal Co-training): Fine-tuned by jointly training on action generation, CoT + subtask generation, and 2D referring expression grounding.
    \item WALL-OSS (Action-only): Fine-tuned using only subtask instructions for action generation.
    \item $\pi_0$ (Action-only): A VLA with a similar architecture to WALL-OSS, fine-tuned only for action generation.
\end{itemize}

As shown in Table~\ref{tab:Instruction_following_acc}, we evaluate each model's ability to select the correct block when provided with a precise instruction (e.g., "pick up the letter A"). The results indicate that the WALL-OSS (Multi-modal Co-training) configuration achieves the highest instruction-following accuracy. While the performance of WALL-OSS (Action-only) declines relative to the multi-modal setup, it remains significantly better than $\pi_0$ (Action-only), whose accuracy is close to random chance.

These findings clearly demonstrate that multi-modal co-training substantially enhances a model’s ability to follow fine-grained instructions. The pre-training phase of WALL-OSS establishes a strong foundation for multimodal action alignment, and maintaining this co-training strategy during fine-tuning further amplifies the model’s instruction-grounded execution capabilities.

\begin{table}[]
\centering
\small
\setlength{\tabcolsep}{4pt}
\begin{tabular}{@{}lccc@{}}
\toprule
\textbf{Block Type} & \textbf{WALL-OSS} & \textbf{WALL-OSS} & \textbf{$\boldsymbol{\pi_0}$} \\
& \textbf{(Co-training)} & \textbf{(Action-only)} & \textbf{(Action-only)} \\
\midrule
Letter                         & 87\%                                                                          & 26\%                                                              & 9\%                                                          \\
Number                         & 95\%                                                                          & 80\%                                                              & 35\%                                                         \\ \bottomrule
\end{tabular}
\caption{Instruction-following accuracy in block-spell task.}
\label{tab:Instruction_following_acc}
\end{table}

\section{Discussion}

\subsection{A Unified Theory of Embodied Foundation Models}
The emergence of foundation models has fundamentally reshaped our understanding of artificial intelligence, yet their extension to embodied domains reveals deeper questions about the nature of intelligence itself: How large are the intrinsic gaps across modalities? Can a unified architecture process heterogeneous modalities without degrading any of them? If so, which approximation trajectories reach that state most efficiently?

WALL-OSS is designed to bridge the language–vision–action gap. Our tightly coupled MoE architecture, multimodal curriculum that augments VLMs with embodied understanding, and multi-stage training schedule collectively yield a unified, flexible, differentiable, end-to-end mapping from high-level instructions—through narrow-sense CoT and subtask decomposition—to discrete and then continuous actions. The rich forward mappings learned during training correspondingly confer flexibility at inference.

Early explorations that aggressively discretized nonlinguistic modalities and trained them jointly with text under a next-token objective proved unsatisfactory. The field subsequently shifted toward conservative side-branches for new modalities, during which original VL priors were heavily disrupted. By contrast, WALL-OSS employs a smaller fraction of additional parameters yet demonstrates that simultaneously improving embodied VL understanding and action generation is both natural and feasible, rather than presupposing a zero-sum trade-off between VL and action capabilities.

Overall, WALL-OSS achieves:  
(i) enhanced multimodal grounding and reasoning ability in embodied environments, thanks to embodied VQA and Uni-CoT design;  
(ii) competitive action generation with stronger OOD generalization than DP, though Pi-0 remains superior for precise manipulation;  
(iii) measurable gains from the Inspiration stage, validating our staged training strategy.  
These findings support our claim that tightly coupled multimodal reasoning and action modeling are key to scalable embodied intelligence.

\subsection{Future Directions}
While unified text-image architectures reach related conclusions, action remains qualitatively distinct: signals are sparse, lie on high-dimensional manifolds, and a compact trajectory corresponds to vast visual and linguistic representations. These representations span high-level state analysis, intent formation, and task decomposition, whereas execution depends on low-level spatial perception and kinematics. Effective modeling must therefore align high-level textual/visual reasoning with low-level perception and control.

Motivated by the former, we introduce large language models; for the latter, we ask whether intermediate formulations (e.g., future-frame prediction) or intermediate modalities (e.g., 3D perception) are needed to reduce the difficulty of learning the VL→A mapping. We view strictly end-to-end VL→A and the intermediate-route approaches as two convergent paths, focusing on their relative efficiency toward AGI.
Video and 3D modalities typically exhibit higher correlation with action than text–action pairs, effectively de-sparsifying supervision; yet 3D data remain scarce. Large-scale video prediction can leverage internet data but introduces substantial redundant supervision—humans act without predicting all details and perceive selectively by intent—prompting bidirectional world-modeling for behavior prediction. With human media largely unchanged, the growth of data and algorithms in each modality is driven by adjacent industries (e.g., AIGC, graphics), which also nurture the "intermediate" priors needed for action modeling. Progress in end-to-end action prediction depends on the embodied industry and likewise amounts to implicit learning of such intermediates. As demand stimulates both academia and industry, the relative efficiency of different routes toward AGI is inherently dynamic.

\clearpage
\newpage
\bibliographystyle{assets/plainnat}
\bibliography{main}

\end{document}